**Intelligent upper-limb exoskeleton integrated with soft wearable bioelectronics and deep-learning for human intention-driven strength augmentation based on sensory feedback**

Jinwoo Lee[1,2]†, Kangkyu Kwon[2,3]†, Ira Soltis[2,4], Jared Matthews[2,4], Yoon-Jae Lee[2,3], Hojoong Kim[2,4], Lissette Romero[2,5], Nathan Zavanelli[2,3], Youngjin Kwon[2,4], Shinjae Kwon[2,4], Jimin Lee[2,4], Yewon Na[2,4], Sung Hoon Lee[2,3], Ki Jun Yu[6], Minoru Shinohara[2,7,8], Frank L. Hammond[4,8], Woon-Hong Yeo[2,4,8,9]*

[1] Department of Mechanical, Robotics, and Energy Engineering, Dongguk University, Seoul, 04620, Republic of Korea.
[2] IEN Center for Human-Centric Interfaces and Engineering, Institute for Electronics and Nanotechnology, Georgia Institute of Technology, Atlanta, GA 30332, USA.
[3] School of Electrical and Computer Engineering, Georgia Institute of Technology, Atlanta, GA 30332, USA.
[4] George W. Woodruff School of Mechanical Engineering, Georgia Institute of Technology, Atlanta, GA 30332, USA.
[5] School of Industrial Design, Georgia Institute of Technology, Atlanta, GA 30332, USA
[6] School of Electrical and Electronic Engineering, Yonsei University, Seoul 03722, Republic of Korea.
[7] School of Biological Sciences, Georgia Institute of Technology, Atlanta, GA 30332, USA
[8] Wallace H. Coulter Department of Biomedical Engineering, Georgia Institute of Technology and Emory University School of Medicine, Atlanta, GA 30332, USA.
[9] Institute for Materials, Parker H. Petit Institute for Bioengineering and Biosciences, Neural Engineering Center, Institute for Robotics and Intelligent Machines, Georgia Institute of Technology, Atlanta, GA 30332, USA.
† J. Lee and K. Kwon equally contributed to this work
*Corresponding author. Email: Dr. Woon-Hong Yeo (whyeo@gatech.edu)

**ABSTRACT**
The age and stroke-associated decline in musculoskeletal strength degrades the ability to perform daily human tasks using the upper extremities. Here, we introduce an intelligent upper-limb exoskeleton system that uses cloud-based deep learning to predict human intention for strength augmentation. The embedded soft wearable sensors provide sensory feedback by collecting real-time muscle signals, which are simultaneously computed to determine the user's intended movement. The cloud-based deep-learning predicts four upper-limb joint motions with an average accuracy of 96.2% at a 500-550 millisecond response rate, suggesting that the exoskeleton operates just by human intention. In addition, an array of soft pneumatics assists the intended movements by providing 897 newton of force while generating displacement of 87 millimeter at maximum. Collectively, the intent-driven exoskeleton can reduce human muscle activities by 3.7 times on average compared to the unassisted exoskeleton.

# INTRODUCTION

Many individuals suffer from neuromotor disorders that primarily arise from stroke-induced and age-associated declines in musculoskeletal strength and control. Statistically, strokes affect one out of four adults over the age of 25 in their lifetime, and 12.2 million of the global population experience stroke each year[1], resulting in neuromotor disorders for 20-40% of victims[2]. The population with neuromotor disorders will amount to a much larger number if we also consider the elderly population with neuromotor disorders. Such a disorder restricts the functional independence of the inflicted population because the reduced motor control and unwanted tremor of the upper limb usually pose considerable difficulties in performing everyday tasks that require the dexterity of the upper limbs. Moreover, neuromotor disorders generate tremendous social expenditure in healthcare. The direct and indirect social costs of stroke amount to approximately $65 billion annually in the United States[3].

There exists an immediate demand to address these problems, and the existing robots are capable of mechanically augmenting human upper-extremity strengths. However, the previously reported exoskeletons cannot provide pragmatic solutions because they lack essential functionalities to augment the upper-extremity movements. Primarily, the most critical limitation of the existing exoskeletons originates from their incapability to predict the real-time intention of the user[4-14]. In other words, many of the existing solutions can only assist the movements of the user as pre-programmed due to the absence of sensory feedback pathways and artificial intelligence to predict the real-time intention of the user. Another limitation of the previously reported exoskeletons is their structural design. The existing exoskeletons use a stationary mode, limiting their potential use to assist everyday tasks in a mobile style[4,10,15-18]. Moreover, these systems rely on complex hardware in large form factors with complicated electrical wiring, which makes it difficult for the users to use and adapt to[6,9,10,12,16-18]. Besides, most of the previous upper-extremity exoskeletons assist only a single joint movement, such as an elbow or shoulder joint movement[4,5,7,8,13,19]. Thus, these systems cannot meet the needs of everyday tasks that use both elbow and shoulder joint motions and combined movements.

In addition, sensory haptic feedback in human assistive robotics is crucial because it translates human physiological signals into strength augmentation. Although a variety of sensory technologies such as strain sensors offered promising features in human motion recognition [20-27], the bodily strain data does not provide much information about human strength. In this context, electromyography (EMG) signals can offer direct information about upper-extremity movements as EMG records the electrical signals in the presence of muscle activities. Still, some existing upper-extremity exoskeletons do not incorporate physiological sensors[4,15,19]. Furthermore, many previous works, including EMG sensors in their systems, use the sensory information to compare the relative muscle involvement rather than to implement the sensory feedback[5,6,8-14]. These critical limitations of the current upper-extremity exoskeletons restrain the realistic use of exoskeletons in daily life. In this respect, we believe that the ideal exoskeleton robot for human strength augmentation should be able to (i) predict the user's intended movement with high accuracy, (ii) be portable, (iii) be lightweight, (iv) be easy to use, (v) support the multiple upper-limb joint movements, and (vi) incorporate the high-fidelity sensory feedback. To the authors' best knowledge, there has been no work that seamlessly integrated all the ideal elements into a fully functional exoskeleton that can assist individuals with neuromotor disorder in performing everyday tasks in a completely user-friendly fashion.

Here, this paper reports a human-intent-driven robotic exoskeleton that assists the upper-extremity joint movements by a deep-learning-based cloud computing platform with high-fidelity sensory feedback using soft bioelectronics. Incorporating the cloud computing platform with the human assistive robot enables the real-time prediction of the user's intended movement with high accuracy. Moreover, an array of pneumatic artificial muscles generates a high magnitude of mechanical force to assist the multiple intended movements of the wearer. Furthermore, we introduce an array of wireless, soft, EMG sensors that can monitor high-quality electrophysiological signals to constitute reliable sensory feedback. Table 1 captures the distinguishable novelty of this work by drawing a comparison between this work and the existing exoskeletons. In addition, Supplementary Table 1 also summarizes the comprehensive comparison with other assistive robots for human upper limb movements that target a similar demographic or purpose.

## RESULTS
**Overview of the system architecture featuring an intelligent upper-limb exoskeleton.**
**Fig. 1** shows the overview of an intelligent upper-limb exoskeleton system that uses cloud-based deep learning to predict human intention for strength augmentation. The embedded soft wearable sensors provide sensory feedback by collecting real-time muscle signals, which are simultaneously computed to determine the user's intended movement. Specifically, the soft sensors attached to the wearer's skin acquire muscle activities through electromyogram (EMG) signals and wirelessly transmit the collected signals to an Android cloud in real time. When the human wearer intends to perform an arbitrary joint movement by activating the associated upper-limb muscles, the multi-channel sensors simultaneously send the EMG signals to the cloud, which conducts the signal processing and the deep-learning algorithm to predict the intended movement. Finally, the prediction output is wirelessly transferred to the exoskeleton such that the robot assists the planned action in real-time. All these sequential processes occur within 350 ms. The wearable exoskeleton system can support four types of upper-limb movements in human strength augmentation: shoulder flexion, shoulder extension, elbow flexion, and elbow extension (**Fig.1**), which are crucial joint movements to perform everyday tasks in the real world. The intelligent robotic system with cloud-based deep learning offers immediate assistance to the user based on the onset of the specific muscle activation. The presented upper-limb exoskeleton integrates soft thin-film sensors and electronics with soft pneumatic artificial muscles (PAMs), packaged as an all-in-one wearable system to maximize user convenience and comfort.

**Design and characterization of PAM and exoskeleton.**
**Fig. 2** delineates the first component of the intent-driven exoskeleton, the PAM and exoskeleton frame, which augment human upper-limb strength for supporting daily tasks. We employed soft pneumatics as a driving force for the system due to its desirable characteristics such as natural compliance, low power consumption, lightweight, high force-to-weight ratio, simple controllability, and inherently safe mechanism[28-30]. Hence, to deliver such benefits of soft pneumatics, we designed PAM that operates just like human muscles, as in **Fig. 2a**, which shows its contraction in the longitudinal direction and generates force as PAM inflates, converting the compressed air energy into mechanical motion. As shown from the exploded view of PAM in **Fig. 2b**, the soft actuator inflates as the air is infused into its elastomeric silicone bladder via the pneumatic tube, and the polyester mesh sleeving serves to restrict the strain of PAM to a certain extent. **Fig. 2c** characterizes the relationship between force and contraction length of a single PAM unit as the air pressure increases from 10 psi to 80 psi. Through maximum strength testing, we found that forces above 900 N can cause the PAM to fail in a controlled manner mechanically. At such high forces, the silicone tubing is cut at the interface between the tubing and aluminum end fittings, which results in the high-pressure air leak out. As a result, the pressure range of the exoskeleton was set from 10 psi to 60 psi, and a 70 psi pressure relief valve was integrated for additional safety. Since the force-contraction profile retains an approximately linear relationship with constant pressure, this relationship can provide the desired exoskeleton assistance force for a given joint angle. It is reported that approximately 10 to 15% of the body weight is needed to lift the arm vertically, so it would require 120 N for a person who weighs 80 kg to lift the arm[31,32]. The result suggests that a single PAM can generate a far greater force for strength augmentation because it can produce 897 N at the pressure of 80 psi, for a PAM with a bladder length of 34 cm, owing to the high force-to-weight ratio of soft pneumatics. A single PAM unit only weighs 104 g, and the exoskeleton system utilizes three PAMs to assist different upper-extremity joint movements. In addition, it demonstrates 87 mm of displacement when pressure of 80 psi is applied to the PAM.

**Fig. 2d** shows the exploded view of the exoskeleton and constituting components. The exoskeleton was constructed primarily out of carbon fiber with machined aluminum connectors and stainless-steel bolts to be strong, stiff, and, more importantly, lightweight (670 g) for the user's comfort. The exoskeleton attaches to the backpack with a super swivel ball joint, which allows the exoskeleton to move naturally with the body. Carbon fiber will weave telescoping tubing, and two telescoping collet sets constitute the main vertical strut between the hip and shoulder joint and the strut between the shoulder and elbow joints. This method allows for easy exoskeleton adjustment to different subject body sizes (**Fig. 2e**). Quick adjustments can be made to the hip-to-shoulder, shoulder-to-elbow, and elbow-to-wrist lengths. However, creating a comfortable interface between the exoskeleton and the user can still present fitment challenges. Nevertheless, by leveraging 3D printing for the upper and lower arm mounts, both "one size fits all," and personalized arm mounts can be easily developed and bolted to the exoskeleton. In this design, the arm mounts are 3D printed as a flat pattern, and hot water thermoformed ergonomically attached to the curved geometry of the arm. An inline ball joint interfaces between the 3D-printed upper arm mount and the carbon fiber tubing for additional arm mobility. We place the PAMs inside the backpack, as in Supplementary Fig. 1, such that PAMs do not interfere with the wearer's natural movements while delivering the required augmentation force via cables as PAMs contract. The backpack also contains compressor and solenoid valves that infuse or remove air into/from each PAM unit rapidly, following the flow diagram and power diagram in Supplementary Fig. 2 and Supplementary Fig. 3, respectively. **Fig. 2f** presents snapshots of the human subject wearing the integrated system of the exoskeleton and backpack. Also, a user wears four skin-like soft sensors to have sensory feedback for the exoskeleton control. The integrated system, which also consists of a battery, cables, a printed circuit board (PCB), and a PAM frame, weighs only 4.7 kg, and the weight of each component is tabulated in Supplementary Table 2. The moderate load should be carryable quickly, even for people with declined strength. The backpack distributes the load evenly across the shoulders and back, making it more stable and comfortable to carry than other methods. The effect of wearing a backpack with a load of up to 10% of body weight remains disputable. However, although wearing the exoskeleton does indeed add extra weight to the user, a large number of studies indicate that 10% of body weight should cause only a minor inconvenience to the body.[33-36] Supplementary Note 1 in the supplementary information describes how 10% of body weight affects the human physiological condition. The PCB in the backpack enables wireless data acquisition from the Android cloud. It provides mechanical force for strength augmentation by controlling the solenoid valves, as illustrated in Supplementary Fig. 4. In addition, the pressure sensor in the PCB monitors the pressure of each PAM in real time such that the user can keep track of the input pressure with the portable device interface.

Supplementary Fig. 5 exhibits the screenshot images of the cloud-based GUI we developed for motion training and informative purposes. As the screenshots show, the user can recognize the real-time pressure information and manually control all the PAMs by moving the toggle bars in the GUI through the portable device simultaneously. Besides, the user can also vent the PAMs if he/she wishes to quit getting assistance from the exoskeleton. In this regard, the real-time monitoring of pressure allows the facile manipulation of PAMs via the GUI and thus enables the user to freely interfere with the strength augmentation process when needed.

**Characterization of soft wearable EMG sensors.**
To incorporate the sensory feedback into the intent-driven exoskeleton, we developed an array of soft wearable sensors that can wirelessly monitor the muscle activity signals with minimized motion artifacts and ultimate user comfort. **Fig. 3a** illustrates the assembled and exploded view of the soft EMG sensor that mainly consists of stretchable gold nanomembrane electrodes, a silicone-based adhesive patch, a flexible circuit, and a customized magnet-chargeable battery that can be switched on and off. The soft sensor developed herein can be attached to the skin surface just as a thin-film adhesive patch due to its extremely small form factor and lightweight nature, as presented in **Fig. 3b**. Unlike the commercial sensor that uses the gel-type rigid electrode, the soft and dry nanomembrane electrode does not cause skin irritation even if it is used for a long-term[37]. We conducted the skin test to validate by wearing the soft EMG sensor and the commercial sensor all day (Supplementary Fig. 6). While the dry nanomembrane electrode in this work did not cause skin irritation, the commercial gel-type electrode left a skin rash due to the inflammatory reaction. We conducted the skin test for the five adult subjects and the subjects were asked to wear both the nanomembrane and commercial electrodes for 12 hours. Furthermore, the skin compatibility of the soft EMG sensor was assessed, considering interaction with sweat (Supplementary Fig. 7), to ensure safety and comfort during prolonged usage. Breathability, essential for prolonged skin contact in wearable systems, was validated in several studies[38,39] utilizing same materials, as demonstrated by moisture vapor transmission rate measurements among commonly used silicone elastomers.

In addition to the inflammation-free nature of the soft EMG sensor, the device makes conformal contact with the deformable skin surface. It thus minimizes the motion artifact mainly due to the serpentine electrode design and compact, flexible wireless circuit. The computational finite element analysis in **Fig. 3c** delineates the readily deformable mechanics of the serpentine nanomembrane. The graphical representation shows the von Mises stress distribution of the serpentine electrode under 30% of strain. It suggests that the stress concentration at the inner circle of the serpentine design enables the electrode to stretch under strain [40]. The experimental study in **Fig. 3d** shows the stretchable mechanics of the electrode, and the result indicates that the normalized resistance remained stable with a slight gradual increase even after 300 stretching cycles under 30% of strain. Besides the serpentine electrode, the flexible circuit used in this work delivers steady wireless EMG data after hundreds of bending cycles despite the presence of the rigid microchips on the board, as in Supplementary Fig. 8. In this regard, the high stretchability of the serpentine nanomembrane and flexibility of the circuit with a miniaturized form factor make a synergistic contribution to the high skin-conformality of the device as it makes intimate contact with the irregular geometry of the various human body surfaces (Supplementary Fig. 9). **Fig. 3e** depicts the circuit design that enables real-time data acquisition of the EMG signals as well as the wireless data transfer to the cloud, thereby eliminating the need for complicated wiring during EMG monitoring. Despite the exceptional strengths of the soft EMG sensor, for the device in this work to be widely utilized in real-life studies, it should demonstrate comparable performance to the commercial EMG sensor. Hence, during repeated elbow flexions, we collected the EMG signals with the EMG sensor in this work and the commercial sensor (BioRadio). **Fig. 3f** explicitly captures a highly equivalent performance of our device because the EMG signals acquired by both devices completely overlap. Besides, as **Fig. 3g** indicates, the soft EMG sensor, which is much smaller and thinner than the rigid sensor, shows a similar quality in signal-to-noise ratio (SNR) compared to the commercial one. Long-term reliability was confirmed by measuring the SNR at six-hour intervals over a day, demonstrating comparable quality to commercial sensors (Supplementary Fig. 10). The reusability of the soft EMG sensor was evaluated by measuring SNR after standard daily cleaning using soap over a period of one week, ensuring consistent performance after each cleaning (Supplementary Fig. 11).

**Control and motion classifications of upper-limb joint movements.**
**Fig. 4a** clarifies how the intent-driven exoskeleton operates based on the flow chart, which comprises three sections: EMG sensing artificial skin, cloud computing, and wearable robotic exoskeleton. It begins with the onset of muscle contraction during an arbitrary upper-limb movement by the user. Next, the multi-channel sensors detect the EMG signals and wirelessly transmit the data to the cloud, which conducts real-time signal processing and motion prediction. Signal processing segments the raw EMG signals into 1.0-second-long signals with a 250 ms overlap and applies a bandpass filter before rectifying the EMG data. Afterward, the deep-learning model based on the convolutional neural network (CNN) and long short-term memory (LSTM) predicts the motion class, which is then wirelessly transmitted to the exoskeleton unit (the detailed information about the deep-learning model is described in the Method section). Based on the predicted motion, the PAMs contract, and the displacement produced by PAM is then translated to the exoskeleton via cables such that the exoskeleton can assist the user's intended movement in real-time. Utilizing cloud computing in our exoskeleton system enables the use of advanced machine learning models such as CNNs and LSTMs, which demand computational resources beyond the scope of local portable devices. This strategy not only secures greater processing capabilities and scalability but also ensures uniform updates and consistent performance across different local devices. Moreover, it allows for real-time data analysis and system adaptability to evolving user patterns without hardware alterations. The centralized computation model also streamlines maintenance and supports the deployment of immediate enhancements without physical intervention on the user's end. **Fig. 4b** shows four soft EMG sensors attached to upper body muscles to collect the muscle activities from the biceps, triceps, medial deltoid, and latissimus dorsi since these muscles are responsible for target upper-limb movements in this work. **Fig. 4c** portrays the collected EMG signals when the human subject advertently contracted the corresponding muscle. All graphs show a substantially decreased amplitude of EMG signals during rest, followed by explosive EMG activities during muscle contractions. In this work, we use the onset signals of muscle activation to determine the user's intended movement to assist the user's movement instantaneously. We devised the logic flow as in **Fig. 4d** to classify and augment four different upper-limb movements in real-time. Starting with the ready state, the intent-driven exoskeleton distinguishes whether the biceps or medial deltoid are activated based on the EMG signals. When the soft EMG sensor detects the onset signal of the biceps, the exoskeleton starts pumping the PAM that augments elbow flexion. The user can pause or continue the movement using the triceps muscle. Thus, if the user activates the triceps muscle, the exoskeleton pauses and returns to the rest position by venting the biceps PAM in the presence of additional triceps activation. We developed this algorithm function because the user might intend to stop moving or want a partial movement instead of a full range of motion. In real-life situations, people do not always use the full range of motions for everyday tasks or even cease to move in the course of the movement. Otherwise, without triceps intervention, the exoskeleton keeps augmenting the elbow flexion. The same pattern applies for shoulder flexion except that the antagonistic muscle is latissimus dorsi. Therefore, the exoskeleton will begin to pump the shoulder PAM when the soft EMG sensor detects the onset EMG signal of the medial deltoid. As with elbow flexion, the shoulder flexion ceases under the latissimus dorsi activation and returns to the rest position if additional activation signals exist. Likewise, the exoskeleton continues shoulder flexion without latissimus dorsi muscle intervention. Furthermore, we can also combine two upper-limb movements to perform a more complex activity. For example, if the user pauses during elbow flexion by triceps activation and then activates the medial deltoid muscle, the exoskeleton augments shoulder flexion, and the same posture can be attained in the reversed order (see dashed arrows in **Fig. 4d**). The combination of upper-limb joint movements makes the exoskeleton much more functional because it allows multiple movements that can be utilized to conduct daily tasks.

In addition, the GUI that we developed in this work offers the cloud-user interface in a way that the user receives visual information about the current status. It also allows the user to control each PAM manually. As illustrated in the first image, the GUI shows whether all the soft EMG sensors and the PAM controlling circuit in the backpack are connected to the cloud. The second image shows the interface during the real-time motion classification, in which the user can see the augmented motion and air pressure for each PAM. The last screenshot depicts the EMG monitoring during the motion training session. Here, the user can pause the ongoing movement assistance of the exoskeleton robot in case of emergency, and the individual PAM can be manually controlled simply by moving the three dots on the GUI in the top right corner.

**Deep-learning model examination and integration into the intent-driven robotic system.**

**Fig. 5a** illustrates the deep learning architecture of each muscle model designed for the classification of muscle activation in our study. The CNN+LSTM model proposed in this work receives epoch-by-epoch 1-second-long filtered EMG signals as inputs, which are then directed to one of the four single models depending on the location of the attached sensor. The individual models classify muscle activation, and their outputs are aggregated to determine the necessary motion activation. Supplementary Table 3, 4, 5, and 6 elaborate on the details of the machine learning layer employed for four muscle activation classifications. To start with, we normalized the filtered EMG signals using the Standard Scaler method to convert them to values ranging from 0 to 1. Then we employed the non-linear activation functions called Leaky Rectified Linear Units (Leaky ReLU). To optimize the CNN architecture, we used ADAM optimization with a learning rate of 0.0001 and utilized the cross-entropy loss function to calculate the error. During optimization, the batch size was set as 20, and a dropout deactivation rate of 0.3 was utilized to avoid overfitting. Early stopping was also used to prevent overfitting by randomly selecting 20% of the data from the training set and using it as a validation set at the beginning of the optimization phase. We conducted learning rate annealing with a factor of 5 once the validation loss stopped improving, and utilized two one-dimensional convolutional layers in the CNN+LSTM model along with four units in the LSTM layers. A single convolutional cell (Conv_1D) includes two convolutional layers, a layer of batch normalization, a Max pooling step layer with a filter size of 4, and a Leaky ReLu function layer. After the data was flattened, it was followed by fully connected layers and passed through a softmax layer to compute the predicted three classes (rest, onset, activation). Through the random selection method, we optimized hyper-parameters, and the training was stopped when two successive decays occurred with no improvement in the network performance on the validation set. We constructed the model architecture based on our previous deep-learning models after major modifications to serve the motion prediction scenario[41,42]. **Fig. 5b** demonstrates real-time upper-limb movement augmentation for four target motions enabled by the intent-driven exoskeleton due to the deep-learning model architecture discussed earlier. For all motions, if the user attempts to perform an arbitrary target motion and activates the corresponding muscle in charge, the exoskeleton detects the onset of the generated EMG signals and instantaneously assists the user's intended movement right after the muscle activation. For instance, if the user at the rest position tries to conduct elbow flexion, the cloud-based deep learning algorithm classifies the motion, and the exoskeleton starts to assist the elbow flexion in real-time, as in Motion 1 of the figure. For Motion 2, the human subject at the rest position activates the triceps muscle. The embedded deep learning algorithm determines the intended movement is the biceps extension, which is then assisted by the robot within a few seconds milliseconds. The same applies to Motion 3 and 4, which deal with joint shoulder movements. Hence, in Motion 3, when the wearer at the rest position attempts to activate the medial deltoid, the exoskeleton quickly captures the user's intention and augments the shoulder flexion. Lastly, the exoskeleton assists the shoulder extension when the user activates the latissimus dorsi, as in Motion 4. Supplementary Movie. 1 also shows the exoskeleton that assists all the upper-limb movements in a continuous real-time sequence along with the GUI screen. In brief, all the user needs to do for strength assistance is to conduct any of the trained upper-limb movements because the system is completely intent-driven. Until now, no work has successfully delivered a fully functional intent-driven prototype with total integration of sensory feedback, strength augmentation, and a human intention predicting algorithm in a way that assists four upper limb movements based on the user's intention in real-time. The CNN+LSTM model developed in this work can classify muscle activation based on filtered EMG signals, and we evaluated its performance using a confusion matrix. We obtained 50 sets of EMG data for each movement data from five human subjects for the model architecture. We used 80% of the obtained EMG data for training and the other 20% of data to construct and validate the deep-learning model. As depicted in **Fig. 5c,** the model achieved a high test accuracy of 95.38% for biceps/triceps muscle activation classification and 97.01% for medial deltoid/latissimus dorsi muscle activation classification The high accuracy (96.2%, on average) achieved in this work corroborates the model's reliability. It suggests that the exoskeleton will assist the joint movement just as intended with minimized errors. The cloud returns motion classes with a 200-250 ms response rate to the tablet in real-time for exoskeleton driver actuation. Considering the 100 ms delay for PAM actuation and the 250 ms delay in the EMG data windows for user intent capture, the overall time for movement assistance is effectively 500-550 ms, which is fast enough to support the joint movement. We expect that the deep-learning model in this work can also be directly implemented in developing advanced prosthetics or assistive devices that can respond to changes in muscle activation in real-time.

**Demonstration of human strength augmentation in real life.**
To examine the quantitative strength augmentation that the intent-driven exoskeleton can provide, we compared the EMG signals when performing the trained upper-extremity joint movements with and without exoskeletons, as in **Fig. 6a** and **b**. **Fig. 6a** demonstrates the EMG signals generated as the human subject repeated elbow flexion with and without the exoskeleton. The result in **Fig. 6b** substantiates substantially reduced muscle activities in the presence of the exoskeleton assistance because the average EMG activation (MVC normalized amplitude) of the biceps with the exoskeleton was 3.9 times lower than that without the exoskeleton during elbow flexion. Here, the human subject repeated the elbow flexion five times, and we plotted mean values for both assisted and unassisted modes along with standard deviations. Similarly, the EMG signals of the medial deltoid exhibited a substantial difference when the exoskeleton augmented the shoulder flexion (**Fig. 6c**). **Fig. 6d** implies that the average MVC normalized signal amplitude was 3.5 times lower with the assistance of the exoskeleton robot, underling the high force and displacement generation capability of the PAM to assist the upper-extremity movements. This experimental study demonstrates that the exoskeleton could serve to reduce the efforts to maneuver the upper-extremity activities significantly. In addition to the strength assistance without any mechanical load, **Fig. 6e** shows the EMG signal reduction even with a moderately heavy load. The human subject was asked to conduct elbow flexion and shoulder flexion while holding the 15 lb. dumbbell. The result shows that the exoskeleton reduced the EMG activities by 1.4 fold and 1.6 fold for elbow flexion and shoulder flexion, respectively (Supplementary Fig. 12), indicating that the exoskeleton can still help reduce muscular activities even when holding considerably heavy objects. In addition to assisting dynamic joint movements, the exoskeleton can help the human subject hold the weight against gravity in a static manner for an extended period of time. For instance, in Supplementary Movie. 2, the human subject was then asked to keep holding the 6.8 kg weight with and without the assistance of the exoskeleton. While the human subject, with the assistance, could lift the weight without much difficulty for longer than 3 minutes, the same human subject could not proceed longer than one minute without the assistance of the exoskeleton. **Fig. 6f** demonstrates the real-life usage examples of the exoskeleton to assist the movement for completing daily tasks. Since the exoskeleton in this work can support a variety of upper-limb joint movements, including the combined movement of shoulder flexion and elbow flexion, it can assist in completing several daily tasks such as placing a box in the shelf, drilling, reaching out for a doorknob, and getting up from a chair with the arm support. Lastly, the cloud-computing platform in this work enables the system to collect massive training data from multiple users who use other local devices with the global data cloud server. It also allows us to update the deep-learning model simultaneously to further enhance the classification accuracy. Table 1 captures the advantages and uniqueness of our exoskeleton compared to prior work. Future work will include establishing the universal, generalized deep-learning model that accurately predicts the intended movement of multiple human subjects, instead of just a single person, such that it can serve to augment the strength of a large mass of human subjects as it is illustrated in Supplementary Fig. 13. In this regard, the cloud-computing platform integrated with the soft bioelectronics and human strength augmenting exoskeleton will not only lay a fundamental groundwork for our future work but also make a substantial contribution to various next-generation human-robot interaction studies.

## DISCUSSION

In summary, this paper reports the total integration of a motion-predicting cloud computing platform, a robotic exoskeleton for strength augmentation, and soft bioelectronics-enabled sensory feedback to develop the fully-intent driven robotic exoskeleton that assists the human upper-extremity joint movements. The PAM module can provide a maximum power source of 895 N of force while generating 87 mm of displacement translated to the exoskeleton frame. As a result, the multiple upper-extremity joint movements can be augmented while retaining a lightweight structure. Soft bioelectronic sensor systems enable sensory feedback by monitoring the EMG signals, which are processed through cloud computing to predict the user's intended movement. The cloud-based deep-learning algorithm can successfully classify four upper-extremity activities with a high test accuracy of 95.4% and 97.0% for biceps/triceps and medial deltoid/latissimus dorsi movements, respectively. The overall response time for the movement assistance, which includes the EMG onset signal detection, cloud-based motion prediction, and actuation, consumes only about 300-350 ms. Moreover, owing to PAM units' high force and displacement generation capability, the intent-driven exoskeleton assists the upper-extremity strength substantially as it reduces the EMG activities by 6.9 and 3.4 times during elbow and shoulder flexion, respectively. In addition, the exoskeleton also assisted the loaded movements by 1.4 ~ 1.6 times for elbow and shoulder flexion, demonstrating that it can be used to carry or move moderately heavy objects in daily life. Overall, this work shows the first demonstration of a class that integrated all the cutting-edge technologies into an intent-driven, fully working robotic exoskeleton that can be directly employed in real-life situations. In this context, we expect that the intent-driven exoskeleton in this study will contribute significantly to next-generation robotics technology and have a transformative impact on the lifestyle of individuals with neuromotor disorders.

## METHODS

**PAM fabrication and characterization.** The actuator consists of a bladder made of ¾" silicone tubing (51135K45, McMaster), a 1 ¼" polyester mesh expandable sleeving (2837K77, McMaster), two custom machined 6061-T6 aluminum end fittings, two ¾" stainless steel PEX-B pinch clamps (Home Depot) and a ¼" push to connect fitting (5779K494, McMaster).

To characterize the PAM, the quasistatic testing of the PAMs was performed with a Mark-10 ESM303 1.5 kN motorized test stand with a Mark-10 Series 5 1 kN force meter (M5-200). The test procedure involved a blocked-force test where the PAMs were held at maximum length and then pressurized. This was followed by programming the motorized test stand to perform a slow unloading and loading ramp at 100 mm min$^{-1}$, where the PAM was allowed to contract until zero force and then stretched back to maximum length. Force and length data were recorded for 10-80 psi in steps of 10 psi.

**Exoskeleton fabrication and integration with the PAM module.** To make the exoskeleton firm, stiff, and lightweight, it was mostly made of carbon fiber with machined aluminum connectors and stainless steel fasteners. The super swivel ball joint (6960T23, McMaster) that connects the exoskeleton to the backpack enables it to move naturally with the body. The main vertical strut between the hip and shoulder joint and the strut between the shoulder and elbow joints were built using two telescoping collet sets (Dragon Plate), 0.5" and 0.625" ID carbon fiber twill weave telescoping tubing. This technique makes it simple to adapt the exoskeleton to various subject body sizes since the lengths from the hip to the shoulder, the shoulder to the elbow, and the elbow to the wrist can all be quickly altered. Fitment problems may arise while attempting to create a comfortable interaction between the exoskeleton and the user. However, by utilizing 3D printing for the upper and lower arm mounts, both universal and custom arm mounts can be quickly created and attached to the exoskeleton. The arm mounts in this design are ergonomically attached to the curved geometry of the arm by being hot water thermoformed after being 3D printed as a flat shape. An inline ball joint (8412K12, McMaster) connects the carbon fiber tubing to the 3D-printed upper arm mount to increase arm movement. As shown in Supplementary Figure 14, 3 mm and 4 mm carbon fiber plates (generic 3K carbon fiber plate, Amazon) are used to construct the shoulder joint. The plate carbon fiber is cut to the appropriate shape using a waterjet, and the plates are sanded and epoxied together (Scotch-Weld™ DP-2216, 3M) to form the joint geometry. A similar manufacturing process is used for the elbow joint. The interface between the telescoping tubing and the joints is machined aluminum and epoxied into the telescoping tubing. To provide smooth joint motions, a standard 605 bearing with an aluminum press-fit housing is bolted into carbon fiber.

**Soft sensor fabrication.** The fabrication process of the proposed system involved three main steps, including fabric substrate fabrication, circuits and encapsulation fabrication, nanomembrane electrode fabrication, and overall assembly. Fabric substrate fabrication involved mixing Silbione (A-4717, Factor II Inc.) parts A and B in a 1:1 weight ratio for ten minutes. The mixed uncured Silbione was then spin-coated on a polytetrafluoroethylene (PTFE) sheet at 1200 RPM for one minute to ensure uniform adhesion layer thickness. The surface was covered with brown fabric medical tape (9907T, 3M) and subjected to curing in an oven at 65°C for 30 minutes. The PTFE sheet was then detached after the Silbione was cured. In the fabrication of circuits and encapsulation, we used a flexible PCB (fPCB) and mounted all electronic components on it with a reflow solder process. Laser cutting was performed to remove unnecessary areas to enhance the mechanical flexibility of the circuit. For power supply and management, we utilized a lithium polymer battery assembly with a slide switch and a circular magnetic recharging port. A low-modulus elastomer (Ecoflex Gel, Smooth-On) was placed underneath the integrated circuit to isolate the strain. The entire electronic system was encapsulated and soft-packaged with an additional elastomer (Ecoflex 00-30, Smooth-On), with only the switch and charging port exposed. The nanomembrane electrode fabrication process involved utilizing gold/chromium electrodes (thickness: 200 nm for Au and 25 nm for Cr), which were deposited by E-beam evaporation and facile laser cutting. PDMS (Sylgard 184, Dow) was employed as the bottom layer of the electrode fabrication due to its proper adhesion and easy-release features. A polymer film (18-0.3F, CS Hyde) was laminated onto the cured PDMS surface, and gold was deposited on the film using an electron-beam deposition process. The film was then laser ablated to obtain a stretchable serpentine pattern of the electrode. Non-functional materials besides the electrode patterns were removed by delaminating them from the PDMS surface with a tweezer. For the final assembly of the electrode system and fabric substrate, gold electrodes were transferred to the fabric's soft adhesive side using water-soluble tape (ASW-35/R-9, Aquasol Corporation). The electrode system-mounted fabric was then patterned using a laser cutting process, and the soft-packaged electronic system was attached to the fabric side of the fabric substrate by adding and curing a thin layer of silicone.

**Soft sensor characterization.** The experimental setup for the mechanical and electrical tests consisted of a digital force gauge (M5-5, Mark-10) and a motorized test stand (ESM303, Mark-10) for measuring mechanical properties, and an LCR meter (Model 891, BK Precision) for measuring electrical resistance. For the cyclic stretching test, the electrode system was stretched and relaxed vertically with a speed of 100 mm/min for 300 cycles, while for the circuit bending test, the system was repeatedly stretched and relaxed at the same speed for 400 cycles. The EMG voltage signal was recorded using a custom android software application (Bio-monitor) during the test. To compare the skin impedance of the electrode, the gold electrode was transferred to the brown tape with the adhesive on it (1200 RPM spin-coated Silbione, Ecoflex 00-30, PDMS). A control group experiment was also conducted with a gel electrode to compare electrode-skin contact impedance. The test was conducted on the biceps, and the electrode attached area was cleaned with a skin preparation gel (NuPrep Skin Prep Gel, Weaver & Co.). The electrode-skin contact impedance was measured by a skin impedance meter (Model 1089NP Checktrode, UFI) connected to two electrodes on the skin. For the signal-to-noise ratio (SNR) calculation of the data, the noise was assumed to be the data measured before elbow flexion without any activity, and its amplitude was computed. The SNR value was calculated using the following equation: $\text{SNR}_{dB} = 20\log_{10}(\frac{A_{\text{Signal}}}{A_{\text{Noise}}})$ Finally, the finite element analysis (FEA) study focused on the mechanical reliability of the two-layered electrode (gold and polyimide) when applied for the yield of 30% strain. We used commercial finite element analysis software (Abaqus, Dassault Systemes Simulia Corporation, Johnston, RI) for simulation and plotted strain distributions with the system on human skin. The simulation used the following material properties: Young's modulus (E) and Poisson's ratio (v): E (PI) = 2.5 MPa and v (PI ) = 0.34 for PI, E (au) = 78 MPa, v (au) = 0.44 for Au.

**Data processing and acquisition.** All data processing was done with Python and the data measured with our system was first processed by a bandpass filter and notch filter. The cut-off frequencies of the bandpass filter used were 10-250Hz for EMG. The stop band frequency of the notch filter was set to 59-61 Hz to remove the power line noise at 60 Hz. During the first stage of EMG acquisition, an instrumentation amplifier (IA) was utilized to amplify the voltage differential between each channel. To prevent electrode DC offset from causing IA saturation, the IA gain was set appropriately. Moreover, the high common mode rejection ratio (CMRR) of the IA reduced the common mode voltage required by the user. A negative feedback circuit utilizing a reference electrode was employed to reduce common mode voltages, which is a popular method for capturing biopotentials. The second stage employed a bandpass filter (BPF) that consists of a low-pass filter and a high-pass filter. The second-order high-pass filter eliminated lower-frequency noise caused by electrode positioning, skin-electrode contact, and other movements. After removing the DC component, the signal was amplified once more, and a potentiometer was used to supply bias voltage to the amplifiers, resulting in a positive output voltage that was adjusted to match the analog-to-digital converter's input signal.

**Data cloud computing interface and firmware development.** The circuit used for wireless recording of non-invasive electromyogram (EMG) signals on the Biceps, Triceps, and Medial Deltoid, Latissimus Dorsi muscles include a microcontroller (nRF52832, Nordic Semiconductor) and a Bluetooth system-on-chip with an analog front-end (ADS1292, Texas Instruments). The ADS1292 IC component and hardware structure are used to organize the analog-to-digital converter (ADC) operation and digital data encoding for Bluetooth buffer transfer. The digital signal with a 10-bit ADC is transmitted with the ADC output's decimal value representing the minimum to maximum muscle movement range. The microcontroller nRF52832 wirelessly transmits data to the mobile device (tablet, Galaxy Tab S8, Samsung) via Bluetooth low energy (BLE). The device is powered by a 40 mAh rechargeable lithium-polymer battery, which has a battery life of approximately 5.1 hours after a full charge. The EMG data from multiple muscle locations are recorded and sent to Google Cloud using a custom Android app on a Galaxy Tab S8 tablet. Cloud computing is used for real-time motion applications due to computing power and processing limitations. The cloud software architecture includes a mobile application that connects captured sensors to cloud data storage, where the machine learning algorithm is hosted. The cloud software preprocesses the data and sends it to a CNN + LSTM algorithm that was developed using the Keres library in Python with other libraries for matrix operations. The cloud returns motion classes with a 200-250 ms response rate to the tablet in real-time for exoskeleton driver actuation. The exoskeleton driver uses a microcontroller (nRF52832, Nordic Semiconductor), internal and external pressure feedback system, and valve control GPIOs. The pressure system provides pressure feedback to the entire system for valve and compressor control. The valve control works with multiple GPIO of the microcontroller, with a response time of 50 ms. The user's onset and specific muscle motion signals, classified from the cloud, actuate the robotic exoskeleton for user strength augmentation.

**Deep-learning motion classifications.** A CNN+LSTM model was utilized in this study using TensorFlow in Python on a laptop equipped with an Intel i7 processor (I7-9750H). The 1-second segmented EMG signals were divided into three parts for model training: 60% for training, 20% for validation, and 20% for testing. The CNN network parameter weight was updated during each training step based on the model training validation accuracy. Hyperparameter values, such as learning rate, kernel size, a filter of each convolutional layer, and the unit of each dropout, were selected using a random search method. The model with the highest validation accuracy was chosen as the best model. To evaluate its performance, the test dataset was used to determine the prediction accuracy of the best model. The study used filtered EMG signals that were normalized using the standard scaler method and Leaky ReLU activation functions. ADAM optimization was used with a learning rate of 0.0001 and the cross-entropy loss function. The batch size was set to 20, and a dropout deactivation rate of 0.3 was utilized to prevent overfitting. Early stopping was also employed with a validation set. The CNN+LSTM model included two one-dimensional convolutional layers, two convolutional cells, and fully connected layers that predicted three classes. Supplementary Table 3, 4, 5, and 6 contain more detailed information about the deep-learning model.

**Strength augmentation demonstration.** To evaluate the effectiveness of the exoskeleton, the EMG signals collected during the experiment were analyzed. Specifically, the average EMG signal amplitude was calculated to quantify the level of muscle activation. The comparison of average EMG signal amplitude between the two conditions (with and without exoskeleton assistance) determined the strength augmentation provided by the exoskeleton. Additionally, the results were compared between the two versions of the exoskeleton (with and without a 10lb dumbbell) to evaluate the impact of the added weight on the effectiveness of the exoskeleton.

**Human subject study.** The study involved 5 volunteers aged 18 or older and the study was conducted by following the approved Institutional Review Board protocol (no. H21214) at Georgia Institute of Technology. Before the in vivo study, all subjects agreed to the study procedures and provided signed consent forms.

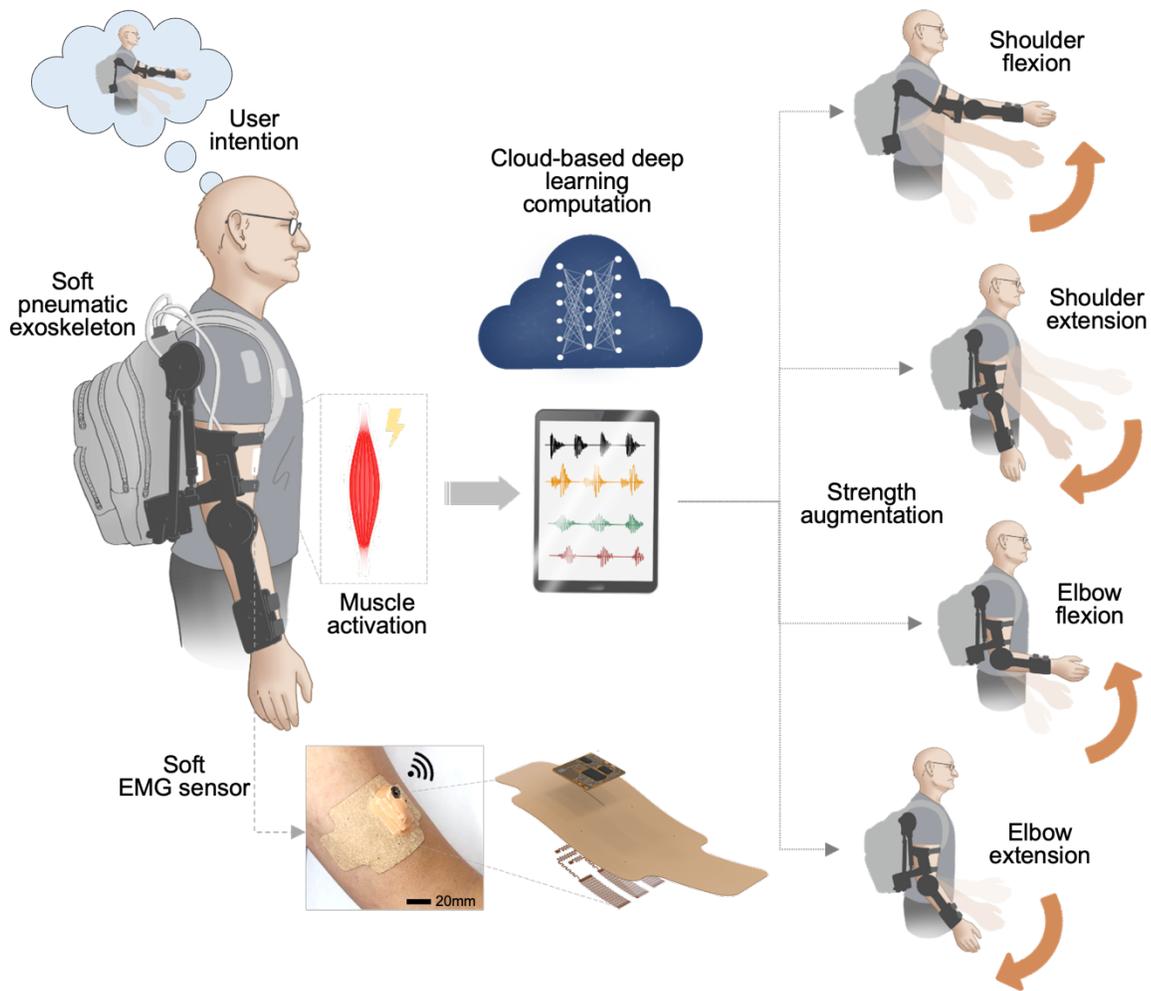

**Figure 1. Overview of the system architecture featuring an intelligent upper-limb exoskeleton with embedded soft sensors and soft actuators**. This system uses cloud-based deep learning to predict human intention for strength augmentation (four different types). A user wears an array of wireless soft EMG sensors on the skin to get sensory feedback. At the same time, the backpack contains three soft pneumatic artificial muscles that operate just as the human muscles contract/relax to translate the generated force and displacement to the exoskeleton frame.

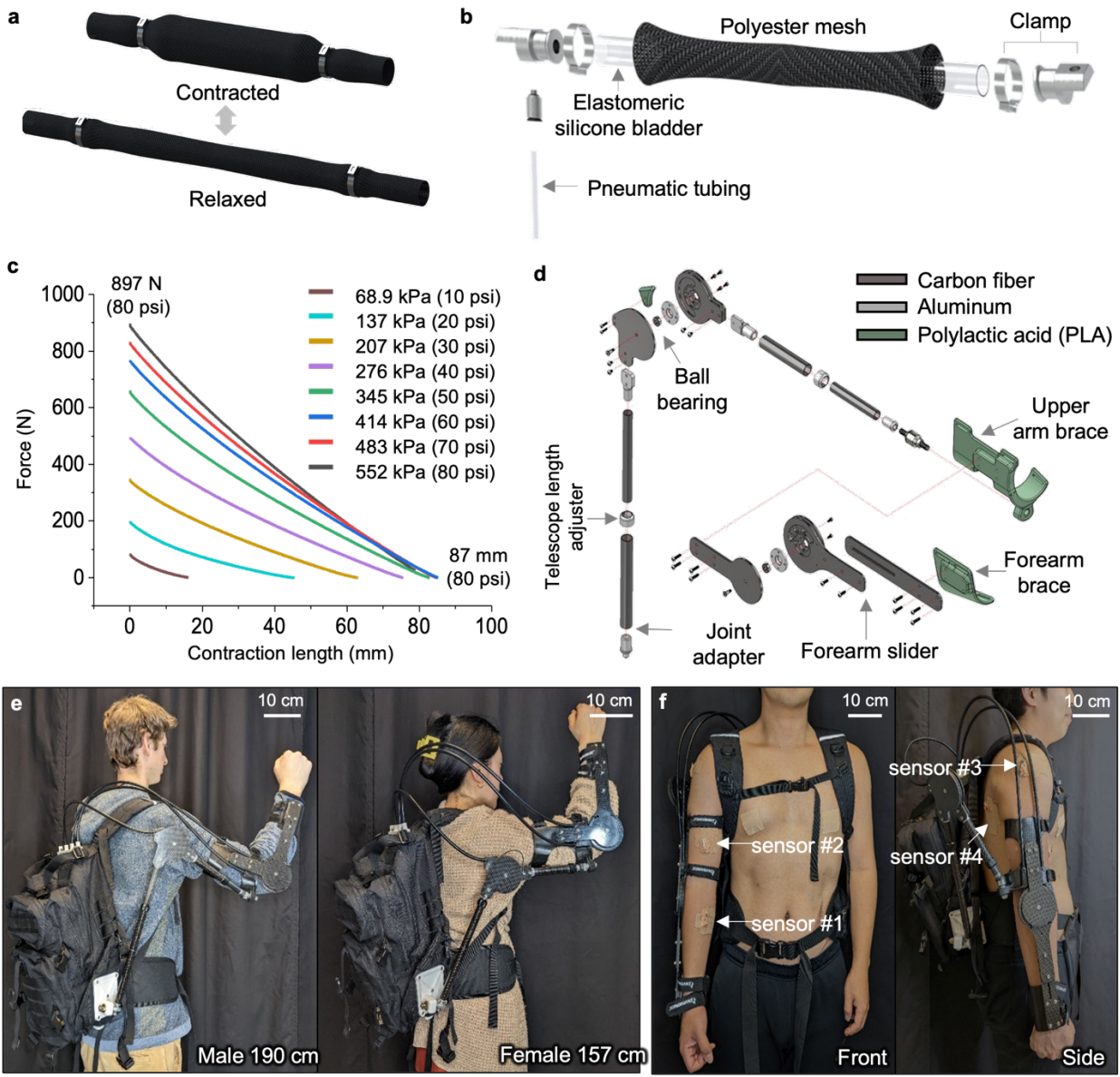

**Fig. 2 Design and characterization of PAM and exoskeleton.** (a) Graphical illustration of the PAM in the contracted and relaxed states. (b) Exploded view of the PAM showing all components. (c) Relationship between force and contraction length as pressure increases from 68.9 kPa to 552 kPa. (d) Exploded view of the exoskeleton frame, including multiple components to cover the forearm and upper arm. The main frame is made of carbon fibers. (e) Photos of two subjects (190 cm male and 157 cm female) wearing the size-adjustable exoskeleton system to capture the "one size fits all" design rationale. (f) Photos of a human subject wearing the exoskeleton and soft sensors. There are four sensors mounted on the skin targeting the biceps, triceps, medial deltoid, and latissimus dorsi.

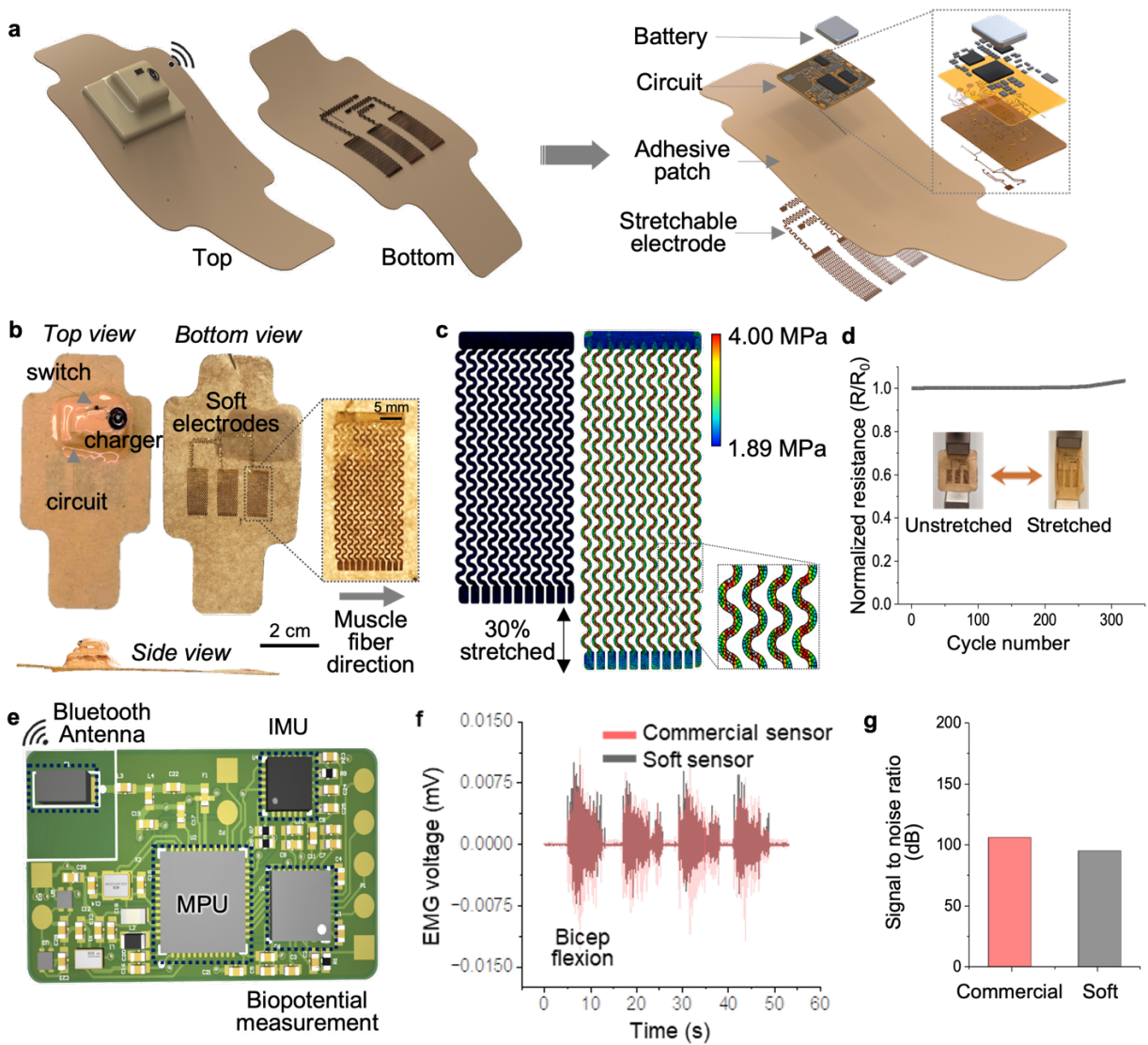

**Fig. 3 Characterization of soft wearable EMG sensors.** (a) Graphical illustration of the soft wireless sensor package (left) and its exploded view (right), showing the stretchable electrode, [43]adhesive patch, battery, and flexible circuits. (b) Top, bottom, and side views of the low-profile ultrathin soft sensor, including a magnetic charging port for multiple uses, a switch to turn the device on and off, and electrodes placed for the muscle fiber direction. (c) Finite element analysis of the stretchable electrode, estimating the mechanical stability upon stretching. The inset figure captures the magnified view of the von Mises stress distribution, which is below the strain limit. (d) Cyclic stretching test of the electrode on the adhesive patch. With 30% tensile strain, the electrode has negligible resistance changes during 300 cycles. (e) Graphical illustration of the wireless circuit that consists of an antenna, microprocessor, inertia measurement unit (IMU), and EMG measuring chip. (f) Comparison of EMG signals from the soft sensor and a commercial sensor during intermittent elbow flexion. (g) Signal-to-noise ratio of the EMG data measured in (f), showing similar measurement quality.

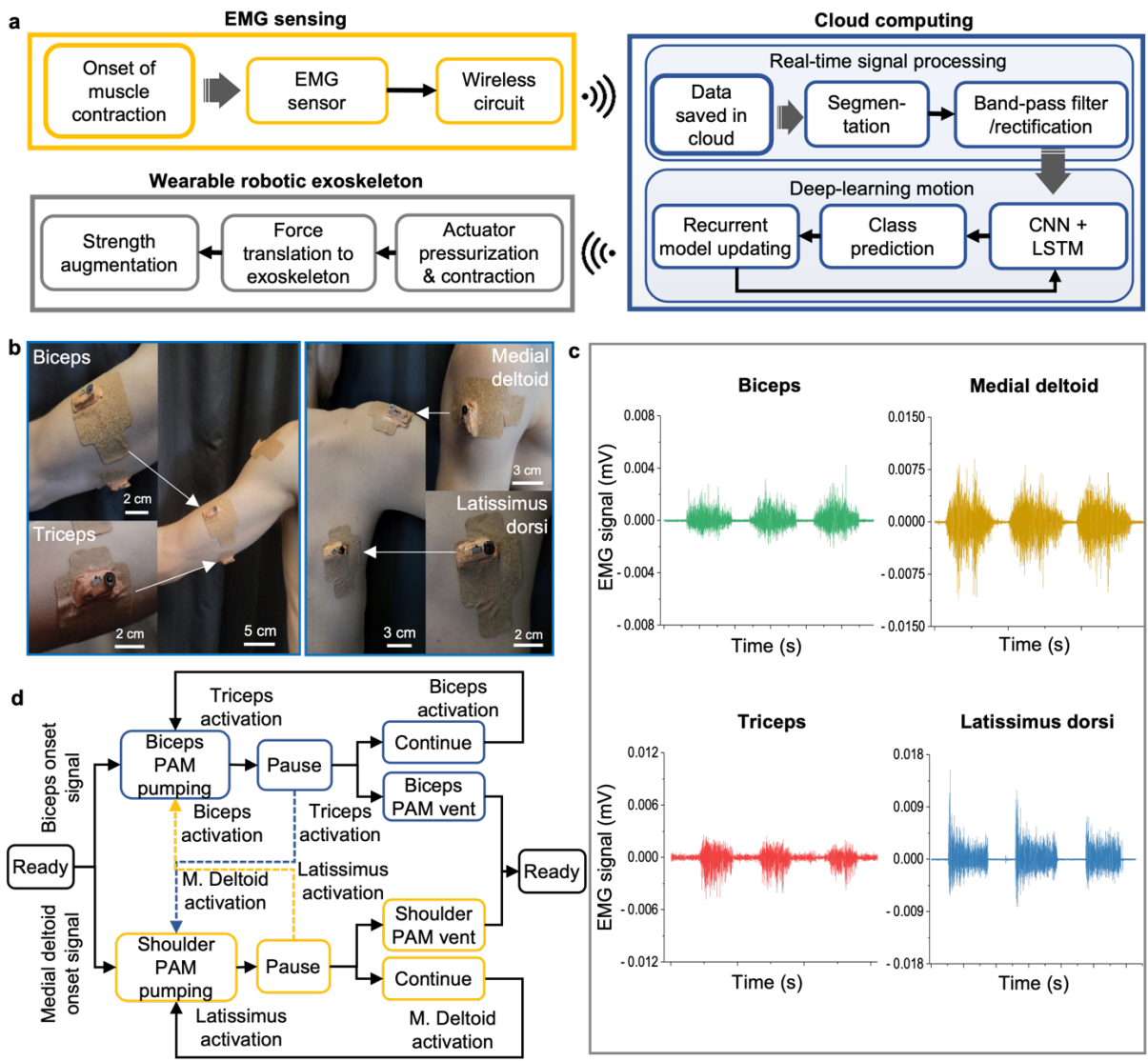

**Fig. 4 Control and motion classifications of upper-limb joint movements.** (a) Flow chart capturing the processes of the intent-driven exoskeleton actuation for strength augmentation, including soft sensor-based EMG sensing, cloud computing, and actuation of soft pneumatics. (b) Photos of skin-mounted soft sensors on a subject. There are four sensors on the skin to record EMG signals from muscles like the biceps brachii, triceps brachii, medial deltoid, and latissimus dorsi. (c) Processed EMG signals measured from four muscles in (b) during intermittent muscle activations. (d) Work logic flow of the upper-limb strength assistance based on sensory feedback, exoskeleton, and deep-learning algorithm for real-time joint movement classification.

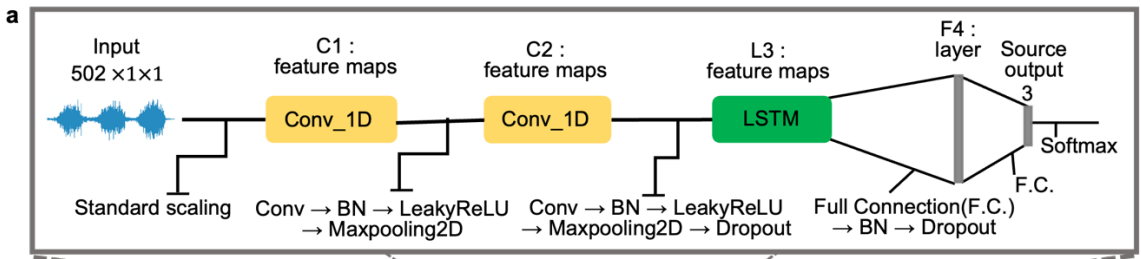
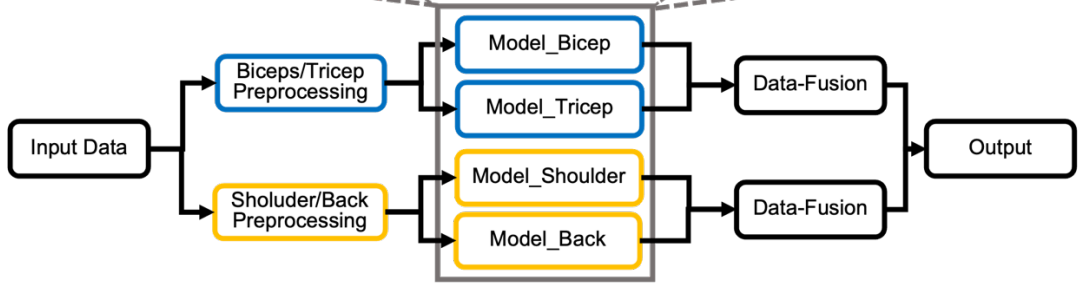
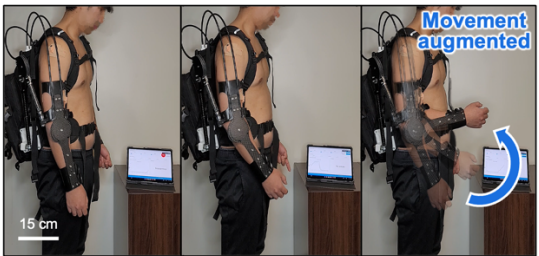
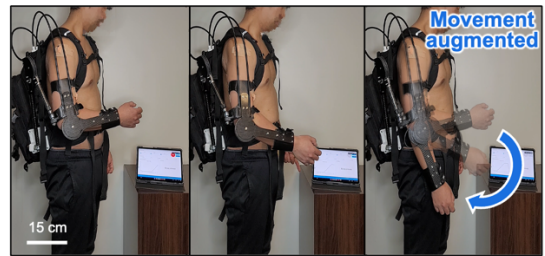
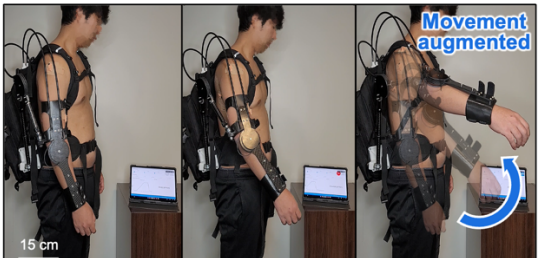
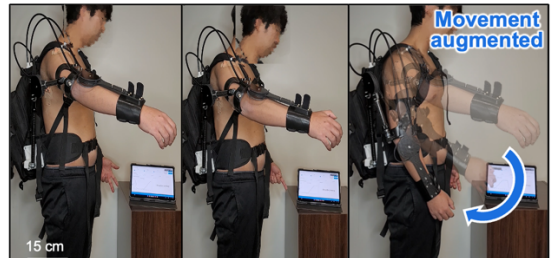
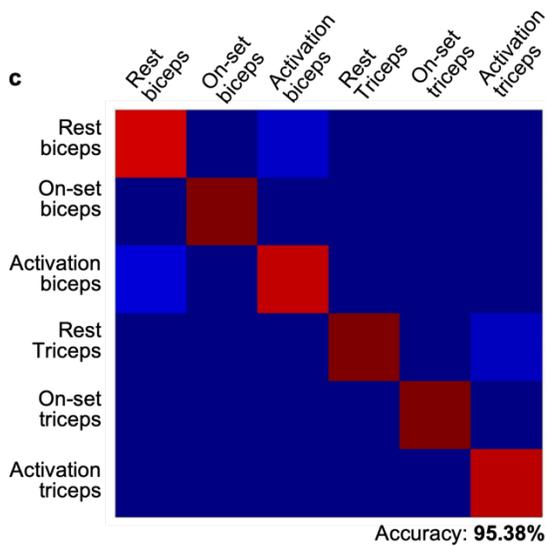
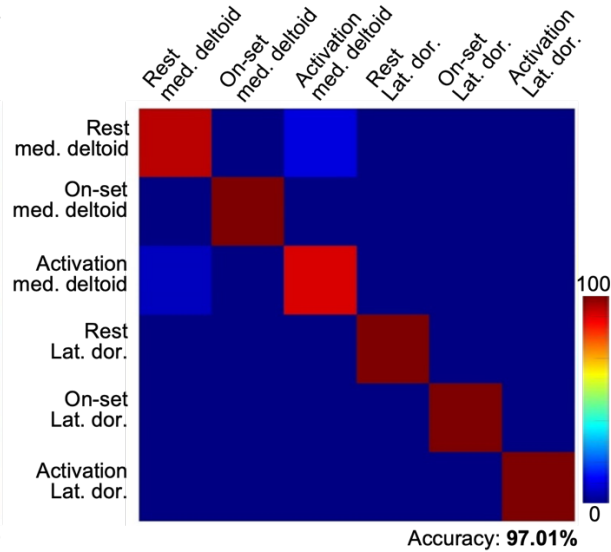

**Fig. 5 Deep-learning model examination and integration into the intent-driven robotic system.** (a) Model architecture of the cloud-based deep-learning algorithm. (b) Demonstration of four augmented motions in real-time in sequential order, which includes elbow flexion, elbowzz extension, shoulder flexion, and shoulder extension. (c) Confusion matrices showing the model classification accuracy of 95.38% and 97.01% for biceps/triceps (left) and medial deltoid/latissimus dorsi (right), respectively.

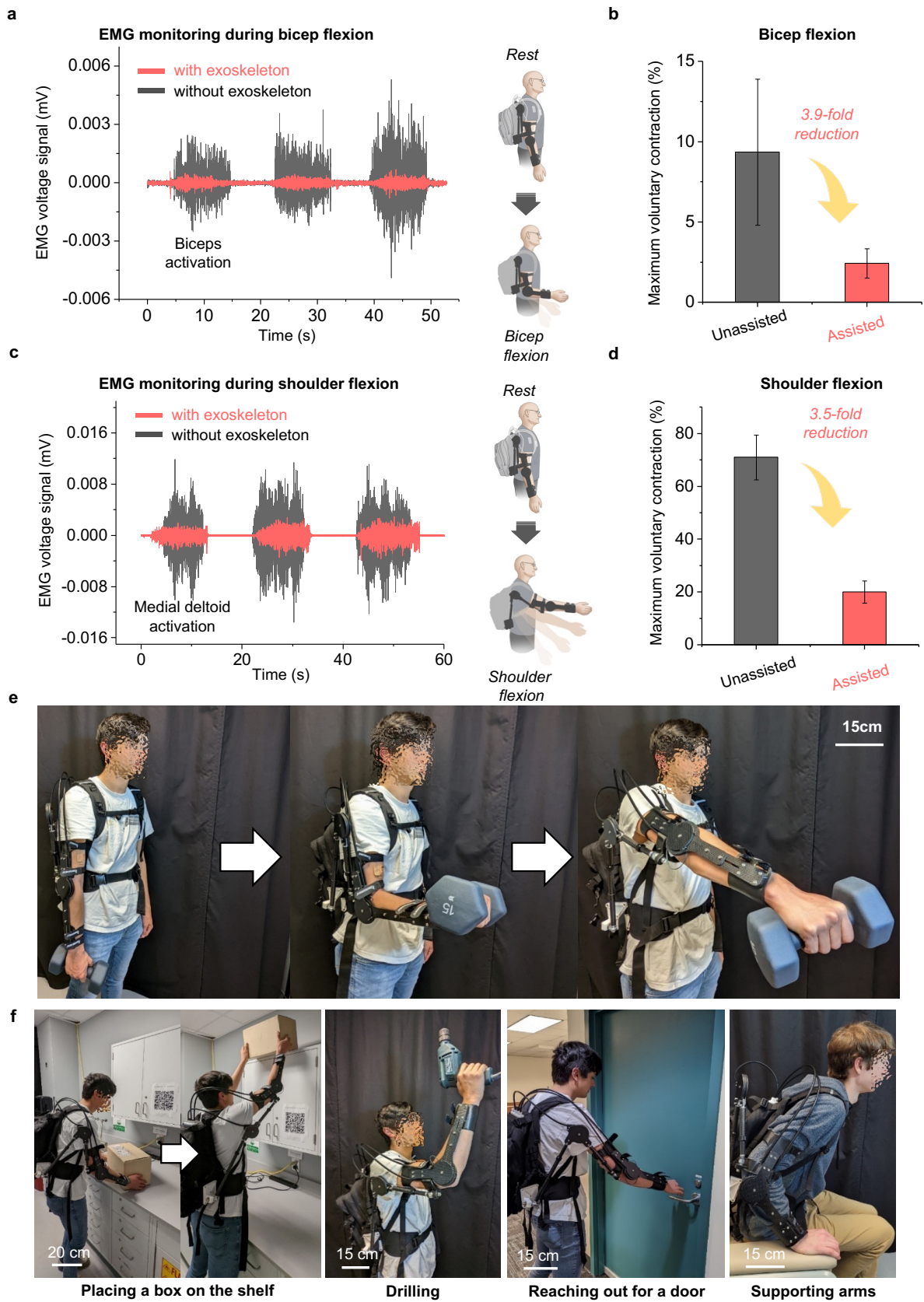

**Figure 6: Demonstration of human strength augmentation in real life.** (a) Comparison of EMG signals between two cases with and without exoskeleton assistance during repeated bicep flexion. (b) Comparison of maximal voluntary contraction (MVC) normalized signals during bicep flexion. This result shows 3.9-fold strength augmentation by the exoskeleton for EMG activation and MVC signals respectively. (c) Comparison

of EMG signals between two cases with and without exoskeleton assistance during shoulder flexion (medial deltoid). (d) Comparison of MVC normalized signals during shoulder flexion. The results indicate that 3.5-fold strength augmentation by the exoskeleton for EMG activation and MVC signals respectively. (e) A subject holding a 6.8 kg weight with exoskeleton assistance, reducing the required power by 1.4 ~ 1.6 times. (f) Potential exoskeleton applications to assist daily activities, such as placing a box on the shelf, drilling, reaching out for a doorknob, and getting up from a chair with arm support.

**Table 1.** Comparison of this work with other upper extremity augmentation exoskeletons.

| References | Integrated system* | Real-time prediction of human intention and constant model update | | Human strength augmentation | | | Sensory feedback | | |
|---|---|---|---|---|---|---|---|---|---|
| | | real-time deep-learning | cloud computing | portable | motions | target movements | sensor configuration | sensor type | wireless |
| This work | O | O | Yes | O | 4 | Shoulder flexion/ extension/elbow extension/ flexion | Skin-conformable | Dry electrode | O |
| 44 | O | O | X | O | 3 | Vertical shoulder flexion/elbow extension/flexion | Rigid | Gel electrode | X |
| 45 | O | O | X | O | 2 | Elbow extension/ flexion | Rigid | Gel electrode | X |
| 16 | O | O | X | X | 2 | Elbow extension/ flexion | Rigid | Gel electrode | X |
| 18 | X | O | X | X | 3 | Shoulder abduction/ Vertical shoulder flexion | Rigid | Gel electrode | O |
| 17 | X | X | X | X | 2 | Elbow extension/ flexion | Rigid | Gel electrode | X |
| 15 | X | X | X | X | 2 | Elbow extension/flexion | No sensor | No sensor | No sensor |
| 7 | X | X | X | O | 1 | Elbow flexion | Rigid | Gel electrode | O |
| 8 | X | X | X | O | 1 | Elbow flexion | Rigid | Gel electrode | X |
| 46 | X | X | X | O | 2 | Elbow extension/ flexion | Rigid | Gel electrode | X |
| 19 | X | X | X | O | 1 | Vertical shoulder flexion | No sensor | No sensor | No sensor |
| 6 | X | X | X | O | 3 | Vertical shoulder flexion/elbow extension/ flexion | Rigid | Gel electrode | X |
| 9 | X | X | X | O | 2 | Vertical shoulder flexion/elbow flexion | Rigid | Gel electrode | X |
| 14 | X | X | X | O | 2 | Shoulder abduction/ Elbow flexion | Rigid | Gel electrode | O |
| 11 | X | X | X | O | 2 | Vertical shoulder flexion/elbow flexion | Rigid | Gel electrode | X |
| 12 | X | X | X | O | 2 | Vertical shoulder flexion/elbow flexion | Rigid | Gel electrode | X |
| 13 | X | X | X | O | 1 | Shoulder abduction | Rigid | Gel electrode | O |
| 5 | X | X | X | X | 1 | Elbow flexion | Rigid | Gel electrode | X |
| 10 | X | X | X | X | 3 | Vertical shoulder flexion/elbow extension/ flexion | Rigid | Gel electrode | X |
| 4 | X | X | X | X | 1 | Elbow flexion | No sensor | No sensor | No sensor |

*Integrated system means the integration of sensory feedback and human intention predicting algorithm into the exoskeleton robotics